\documentclass[11pt]{article}

\usepackage[final]{acl}

\usepackage{times}
\usepackage{latexsym}

\usepackage[T1]{fontenc}

\usepackage[utf8]{inputenc}

\usepackage{microtype}
\usepackage[table]{xcolor}

\usepackage{inconsolata}

\usepackage{graphicx}
\usepackage{tikz}
\usepackage{pgfplots}
\usepgfplotslibrary{groupplots}

\usepackage{amsmath}
\usepackage{amsmath,amssymb}
\usepackage{multirow}
\usepackage{booktabs}
\usepackage{tcolorbox} 
\tcbuselibrary{breakable}
\usepackage{arydshln}

\usepackage{tikz}
\usepackage{pgfplots}
\usepackage[table]{xcolor} 
\usepackage{colortbl}

\definecolor{mygray}{gray}{0.9} 
\definecolor{mylightblue}{rgb}{0.8, 0.9, 1.0} 
\title{SERM: Self-Evolving Relevance Model with Agent-Driven Learning \\ from Massive Query Streams}

\author{
Chenglong Wang\textsuperscript{1}\thanks{Authors contributed equally.}\thanks{This work was done while interning at ByteDance.}, 
Canjia Li\textsuperscript{2}$^*$, 
Xingzhao Zhu\textsuperscript{3}$^\dagger$, 
Yifu Huo\textsuperscript{1}$^\dagger$,
Huiyu Wang\textsuperscript{2},
Weixiong Lin\textsuperscript{2}, \\
\textbf{
Yun Yang\textsuperscript{2},
Qiaozhi He\textsuperscript{2},
Tianhua Zhou\textsuperscript{2},
Xiaojia Chang\textsuperscript{2},
Jingbo Zhu\textsuperscript{1}
and Tong Xiao\textsuperscript{1}\thanks{Corresponding author.}} \\
\textsuperscript{1}Northeastern University, Shenyang, China \quad
\textsuperscript{2}ByteDance \quad \textsuperscript{3}Peking University \\
\ttfamily{clwang1119@gmail.com \quad licanjia@bytedance.com} \quad
\ttfamily{xiaotong@mail.neu.edu.cn}
}

\begin{document}
\maketitle
\begin{abstract}
Due to the dynamically evolving nature of real-world query streams, relevance models struggle to generalize to practical search scenarios.
A sophisticated solution is self-evolution techniques. However, in large-scale industrial settings with massive query streams, this technique faces two challenges: (1) informative samples are often sparse and difficult to identify, and (2) pseudo-labels generated by the current model could be unreliable.
To address these challenges, in this work, we propose a \underline{\textbf{S}}elf-\underline{\textbf{E}}volving \underline{\textbf{R}}elevance \underline{\textbf{M}}odel approach (SERM), which comprises two complementary multi-agent modules: a \textit{multi-agent sample miner}, designed to detect distributional shifts and identify informative training samples, and a \textit{multi-agent relevance annotator}, which provides reliable labels through a two-level agreement framework.
We evaluate SERM on a large-scale industrial platform, which serves billions of user requests daily.
Experimental results demonstrate that SERM can achieve significant performance gains through iterative self-evolution, as validated by extensive offline multilingual evaluations and online testing.

\end{abstract}

\section{Introduction}
Search relevance is central to modern information retrieval, aiming to rank documents that best satisfy a user query \cite{yin2016ranking,li2015toward}. With the explosion of information on platforms such as Google and TikTok, effective relevance modeling has become increasingly important \cite{chen2024towards}. Traditional approaches encode queries and documents into vectors and learn a scoring function \cite{gao2020understanding,zou2021pre}, while recent work leverages large language models (LLMs) to directly generate relevance judgments \cite{zhuang2024setwise,ye2025applying}.

Despite recent progress, relevance modeling still suffers from significant generalization limitations. This stems from the dynamic and continuously evolving nature of real-world query distributions, which makes it difficult for relevance models to generalize effectively to practical search scenarios.
For instance, on online search platforms, users often issue queries containing newly emerged expressions or cultural references, such as \textit{remember me pets arriving on 10/27}. Such queries encode nuanced meanings that models often fail to capture, leading to mismatches between user intent (i.e., \textit{commemorating deceased pets}) and retrieved content (i.e., \textit{generic pets returning home}).

There is much work on addressing this limitation by pre-training models on large-scale document corpora \cite{zou2021pre,zhang2023two,ma2024fine}. However, such approaches primarily capture domain-specific knowledge from static data and fail to account for the fact that real-world query streams continuously introduce novel expressions and emerging linguistic patterns. A more sophisticated approach is self-evolution techniques, including self-training \cite{gulcehre2023reinforced} and self-reflection \cite{huang2022large}, where a model leverages its own predictions on unlabeled user queries to enhance its generalization. While such approaches have shown promise, applying self-evolution in industrial-scale search scenarios poses two fundamental challenges: (\textbf{C1}) informative samples are sparse and difficult to identify within massive query streams, and (\textbf{C2}) pseudo-labels generated by the current model may be unreliable, potentially leading to error accumulation.

To address these challenges, we explore strategies for self-evolving relevance models that can continuously adapt to massive query streams.
To this end, we develop an LLM-based relevance model that, given a query and a document, generates both a relevance score (e.g., 0–4) and the corresponding rationale. Building on this foundation, in this work, we propose a \underline{\textbf{S}}elf-\underline{\textbf{E}}volving \underline{\textbf{R}}elevance \underline{\textbf{M}}odel approach (SERM), which comprises two complementary multi-agent modules: a \textit{multi-agent sample miner} and a \textit{multi-agent relevance annotator}. Specifically, the sample miner monitors incoming queries, detects distributional shifts driven by diverse user behaviors, and selects informative training samples where the model lacks sufficient knowledge to make accurate predictions (tackling \textbf{C1}). The annotator then produces reliable learning signals for these samples through a two-level agreement framework, enabling the model to iteratively refine itself and adapt to emerging user intents (tackling \textbf{C2}). To the best of our knowledge, we are the first to investigate the self-evolution of relevance models using massive query streams.

We employ a large-scale industrial search platform as a testbed to evaluate the effectiveness of our approach.
Specifically, we conduct experiments on massive query streams to demonstrate how our SERM approach enables continuous model evolution.
The experimental results demonstrate that SERM effectively enables the self-evolution of relevance models at scale with massive query streams. Notably, after three iterations of self-evolution, SERM achieves a +2.99 point improvement in NDCG@1 over the baseline model. Moreover, unlike self-training, which relies solely on the model’s own predictions and often suffers from severe error propagation, SERM provides reliable labels that mitigate this issue, enabling consistent and sustained performance gains.



\section{Preliminaries}
\subsection{Task Formulation}
Given a query $q$ and a collection of candidate documents $D=\{d_1, d_2, \cdots, d_m\}$, where $m$ denotes the number of documents, the goal of a relevance model is to compute a score for each document \cite{macavaney2019cedr,nogueira2019multi}. These scores are used to rank documents, allowing the system to return results that best match the query intent. Here, we focus on document search as a representative case of relevance modeling, though the term ``document'' can also encompass other searchable content such as videos or images.

\begin{figure*}
    \centering
    \includegraphics[width=0.99\linewidth]{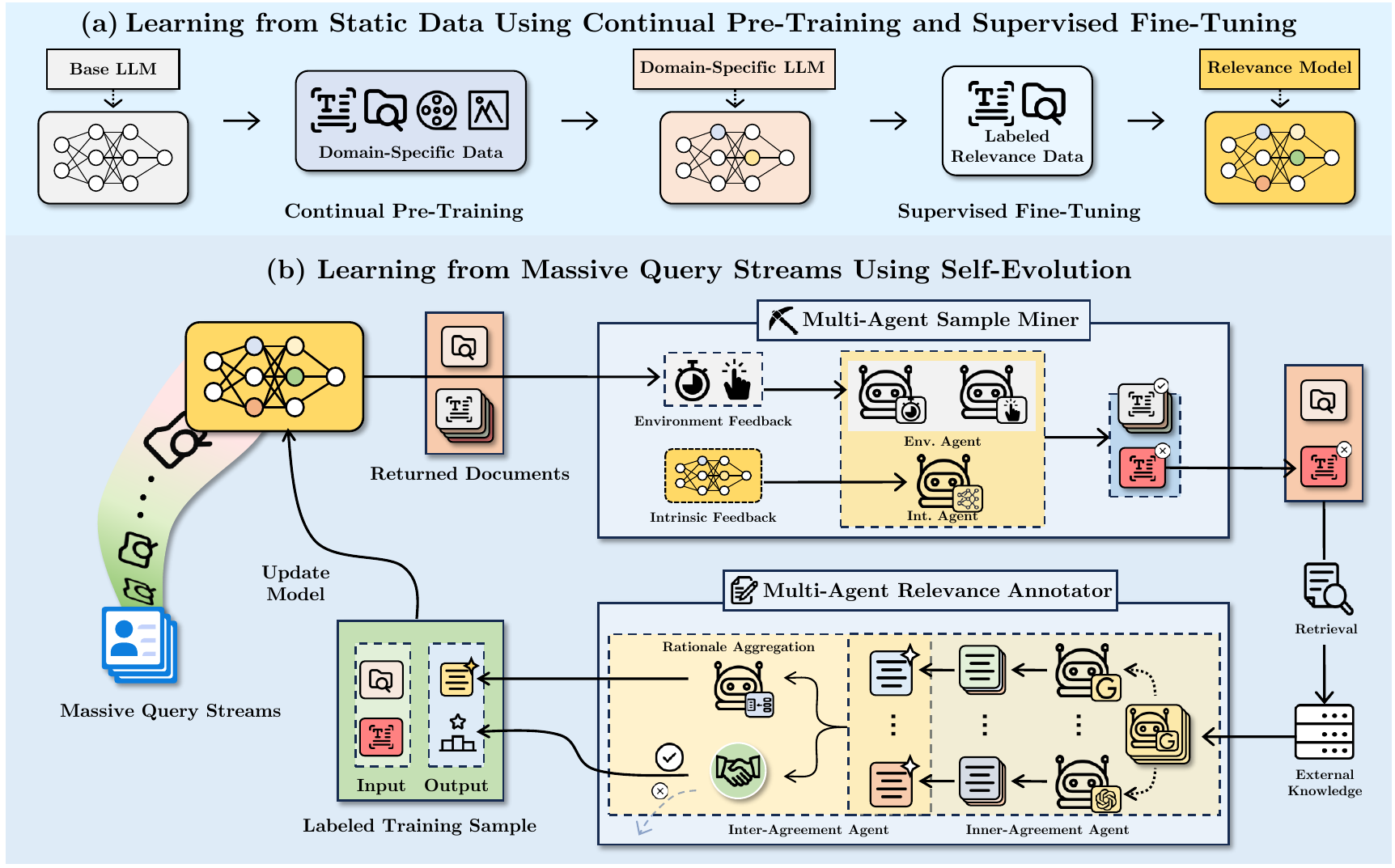}
    \vspace{-2mm}
    \caption{
    \textbf{(a) Learning from Static Data:} The conventional training recipe is to first apply continual pre-training and then perform supervised fine-tuning on static labeled data.
    \textbf{(b) Learning from Massive Query Streams:} The proposed SERM uses a multi-agent sample miner to identify informative samples and a multi-agent relevance annotator to generate reliable labels, enabling continuous model evolution with massive query streams.
    }
    \vspace{-4mm}
    \label{fig:main_figure}
\end{figure*}

\subsection{Continual Pre-Training}
As shown in Figure \ref{fig:main_figure}(a), the development of search relevance models typically follows a two-stage recipe: continual pre-training followed by supervised fine-tuning (SFT). 
There are two commonly used approaches for continual pre-training. One simple approach is to pre-train the model on a large corpus of documents. This allows the model to better capture domain-specific semantics and patterns \cite{zou2022pre,wu2024cprm}.

A second approach is to model relationships between document attributes. For instance, before supervised training, we can generate tasks where the model predicts certain document attributes like post content, thereby encouraging the model to understand the internal relationships between different parts of a document \cite{zhang2023two}. 

\subsection{Supervised Fine-Tuning}
After continual pre-training, we use labeled data to train the model further through SFT. At this stage, modeling approaches can be broadly categorized into \textit{discriminative} and \textit{generative} methods.
\paragraph{Discriminative Relevance Modeling.}
The discriminative modeling approach uses a pre-trained encoder to represent the query $q$ and document $d$ as feature vectors, which are then fed into a scoring function to produce a relevance score. Training typically follows either a pairwise ranking objective or a regression-based objective. 
For the pairwise ranking objective, given a non-relevant document $d_{a}$, the model is trained to assign a higher score to $d_{b}$ than to $d_{a}$ through a Bradley–Terry loss function \cite{bradley1952rank}.
\begin{eqnarray}
\mathcal{L}_{\text{d}}
&=&-  \mathbb{E}_{(q,d_{a},d_{b}) \sim \mathcal{D}_r} \nonumber \\ 
&&\ \ \ \Big[\log \sigma\big(f(q,d_{b}) - f(q,d_{a})\big)\Big]
\end{eqnarray}
where $D_{r}={(q, d_{a}, d_{b})}$ denotes the labeled pairwise dataset, $f(\cdot)$ is the relevance scoring model, and $\sigma(\cdot)$ is the sigmoid function. 
Additionally, when graded relevance labels $y\in\mathcal{Y}$ are available, where $\mathcal{Y}$ denotes the set of label tokens (e.g., $\{0,1,2,3\}$), a regression objective can also be adopted by minimizing the mean squared error between the predicted score and the label.

\paragraph{Generative Relevance Modeling.}
The generative modeling approach utilizes the generation capability of LLMs to directly produce relevance judgments.  
Instead of encoding the query and document into feature vectors and training a scoring function, the model is trained to generate a discrete relevance label.  
The training objective is formulated as a cross-entropy loss over the target label tokens:
\begin{eqnarray}
\mathcal{L}_{\text{g}}
&=& - \mathbb{E}_{(q,d,y) \sim \mathcal{D}_r} \log \mathrm{Pr}_\theta(y | q,d)
\label{eq:generative_modeling}
\end{eqnarray}
where $\mathrm{Pr}_\theta(\cdot)$ is the probability distribution defined by an LLM with parameters $\theta$. This modeling approach enables the model not only to predict a relevance score but also to generate interpretable rationales before the score, thereby enhancing robustness in dynamic search scenarios \cite{zhuang2024setwise,ji2025reason}. Once trained, the model can be used as $f(\cdot)$ as described in Eq. \ref{eq:obtain_score_from_grm}.

\section{Self-Evolving Search Relevance Models}
Our goal is to achieve self-evolution, enabling the model to continuously adapt and refine its relevance predictions based on evolving user queries and behaviors. To achieve this, we propose the SERM, which consists of two multi-agent modules: the multi-agent sample miner (MSM) and the multi-agent relevance annotator (MRA), as illustrated in Figure \ref{fig:main_figure}(b). The following subsections describe these modules in detail.

\subsection{Multi-Agent Sample Miner}
The MSM is designed to identify informative and challenging query–document pairs from massive query streams to drive the self-evolution of the relevance model. The basic idea is that not all samples contribute equally to model improvement: easy or well-predicted pairs provide little additional information, whereas those that reveal the model’s weaknesses are far more valuable for further learning. To this end, the MSM employs multiple complementary agents, as described below.

\subsubsection{Environmental Feedback Agents}  
We design two agents that capture inconsistencies between the relevance model and its surrounding environment.  
These two agents use feedback signals from user interactions and auxiliary models to identify query–document pairs that are potentially informative and challenging for further learning:

\paragraph{User Feedback Agent.}  
We design an agent that employs user interaction signals, including clicks and dwell time, to identify informative query–document pairs for training relevance models. This agent evaluates each pair $(q, d)$ based on two criteria.
First, it detects strong positive user engagement, determined by $\big(\mathbf{1}{\text{click}}(q, d) = 1 \big) \lor \big(U(q, d) > \tau_{u}\big)$, where $\mathbf{1}{\text{click}}(\cdot)$ indicates whether document $d$ was clicked for $q$, $U(q, d)$ is a dwell-time–based engagement metric, and $\tau_{u}$ is the engagement threshold.
Second, it assesses the pair’s difficulty for the current relevance model by checking whether $f(q, d) < \tau_{c}$, where $\tau_{c}$ is the confidence threshold.
When both conditions are met, the agent outputs a feedback signal highlighting the discrepancy between strong user interest and low model confidence, thereby recommending the pair as a valuable candidate for guiding self-evolution.

\paragraph{Click Model Feedback Agent.}
We design a click model feedback agent that augments user feedback by compensating for the biases and sparsity of raw click signals.
In practice, user clicks are often influenced by position and presentation biases \cite{bar2009presentation} and tend to be sparse or delayed, particularly for tail queries or newly introduced documents \cite{chuklin2022click}.
To mitigate these issues, the agent leverages a pre-trained click model $R_{\mathrm{cm}}$, which estimates the probability that a user would click a candidate document given a query. Instead of relying solely on raw click logs, the agent evaluates each query–document pair by comparing the predicted click probability from $R_{\mathrm{cm}}$ with the relevance model’s confidence. If the predicted click probability exceeds a threshold $\tau_{\mathrm{cm}}$, the agent treats the pair as clicked, i.e., $\mathbf{1}_{\text{click}}(q,d)=1$.

\subsubsection{Intrinsic Feedback Agent}
\label{sec:intrinsic_feedback_agent}
The intrinsic feedback agent uses internal signals of the relevance model to identify query–document pairs that best reveal its weaknesses.
Specifically, we design two feedback signals: model disagreement and model uncertainty. 
\paragraph{Model Disagreement.}
We prompt the LLM to generate $K$ relevance judgments with accompanying rationales using temperature sampling, producing a set of scores $\{f^{k}(q, d) \}_{k=1}^{K}$.
The disagreement among these judgments is quantified as
\begin{eqnarray}
\mathrm{MD}(q,d) = \max_{i,j} \big| f^{i}(q,d) - f^{j}(q,d) \big|
\end{eqnarray}
where a larger value indicates greater inconsistency in the model’s predictions.

\paragraph{Model Uncertainty.}
We compute the model’s prediction uncertainty using the entropy of the relevance label distribution:
\begin{eqnarray}
    \mathrm{MU}(q,d) &=& - \sum_{y} \mathrm{Pr}_{\theta}(y \mid q,d) \nonumber \\
    &&\times \log \mathrm{Pr}_{\theta}(y \mid q,d)
    \vspace{-2mm}
\end{eqnarray}
where higher entropy denotes lower confidence and suggests that the pair is harder for the model. By jointly considering both disagreement and uncertainty, the agent identifies samples with high inconsistency and low confidence, which are particularly informative for guiding the model’s self-evolution.

Using the designed agents, the MQM module selects query–document pairs that meet the specified conditions. Specifically, for each new query, it samples $n$ candidate documents identified by each agent, forming a set of trainable query–document pairs. Duplicate documents are removed during this process. The union of these sampled pairs constitutes a collection of hard and informative cases, which are then passed to the MRA module to generate reliable learning signals.

\subsection{Multi-Agent Relevance Annotator}
Building on the samples identified by the MSM, we propose a two-level agreement framework to enhance the accuracy and robustness of automated relevance annotation using multiple large-scale models. This framework operates in two stages: inner-agreement and inter-agreement, ensuring that the selected query–document pairs are annotated with both reliable labels and coherent rationales.

\paragraph{Inner-Agreement Agent.}
The inner-agreement agent operates by leveraging multiple LLMs (e.g., GPT-4o and Gemini2.5-Pro) to perform relevance annotation. Acting as autonomous evaluators, the agent first retrieves external knowledge relevant to the given query–document pair and then reasons about its relevance using the retrieved context. To ensure robustness, the agent employs a multi-path chain-of-thought strategy \cite{thomas2024large}, generating multiple independent reasoning paths for each pair. These reasoning paths produce a set of candidate relevance labels, which the agent consolidates via majority voting to arrive at a stable label prediction. By internally reconciling diverse reasoning paths, the inner-agreement agent mitigates the randomness of single-pass LLM outputs and improves intra-agent reliability.

\paragraph{Inter-Agreement Agent.}
The inter-agreement agent operates by reconciling the outputs of multiple inner-agreement agents. After receiving stable label predictions from each agent, it filters the results by retaining only those query–document pairs on which the agents reach consensus, thereby constructing a high-confidence annotated dataset. Beyond label selection, the inter-agreement agent also assumes responsibility for rationale generation. To this end, it collects all reasoning paths that support the agreed-upon relevance label and consolidates them into a single, coherent explanation. By enforcing cross-agent consensus and producing unified rationales, the inter-agreement agent enhances annotation quality and ensures that only reliable signals are passed forward for self-evolution.

Importantly, although MRA employs external feedback, it fundamentally differs from knowledge distillation \cite{kim2016sequence}. A detailed discussion of these differences is shown in Appendix~\ref{app:diff_from_kd}. In essence, the MRA module functions as a form of evolutionary feedback that iteratively corrects and refines the model's outputs, rather than transferring static knowledge from a fixed teacher. This design aligns with recent work on self-evolving systems, where external tools or auxiliary models are employed to refine model predictions and provide feedback signals for sustained self-improvement \cite{goucritic,zhou2024isr}.

\section{Experiments}
We evaluate our self-evolving approach on an online social platform, focusing on a document search task and employing the widely used Qwen2.5-7B and Qwen2.5-1.5B models. Notably, the platform handles a massive daily volume of user queries from multiple countries, offering a realistic and challenging testbed for our research problem.

\subsection{Datasets}
For continual pre-training, we used a corpus of 100B tokens, primarily derived from the platform’s document collection.
For SFT, we employed 3.6M labeled query–document pairs, each labeled according to its relevance score (ranging from 0 to 3, indicating bad, fair, good, and excellent, respectively).
During self-evolution, each iteration sampled approximately 700K user query–document pairs from the platform’s database to run SERM, with iterations spaced two weeks apart to ensure sufficient shifts in the user query distribution.
Note that since the platform serves users across multiple countries, our datasets were multilingual, with the language distribution shown in Table \ref{tab:data_dist} in the Appendix.
Additionally, given that this work focuses on evolving relevance models from massive real-world user queries, existing open-source datasets did not meet our research needs. Consequently, all datasets used in this study were collected from real-world industrial application scenarios.

\begin{table*}[!t]
    \centering
    \resizebox{\linewidth}{!}{
    \setlength{\tabcolsep}{6.5pt}
\begin{tabular}{lccccccccc}
\toprule[1.1pt]
\multirow{2}{*}{Method} &   \multicolumn{3}{c}{Germanic} & \multicolumn{3}{c}{Romance} & \multicolumn{3}{c}{Minor Language} \\ \cmidrule(r){2-4} \cmidrule(r){5-7} \cmidrule(r){8-10}
&ND@1 & ND@4  & Acc.  & ND@1     &  ND@4  &Acc.  &ND@1 &ND@4  &Acc.  \\ \midrule
\multicolumn{5}{l}{\textbf{\textit{Training with Qwen2.5-7B Model}}}\\ \midrule
CT+SFT & 84.74 & 86.68 & 55.30 & 85.61 & 87.33 & 50.74 & 82.02 & 84.02 & 52.89  \\ 
\midrule
Self-Training  \\
\ \ \ \ \ \textit{Iteration 1}  & 84.87 & 86.75 & 55.74  & 85.82 & 87.43 & 51.40  & 82.30  & 84.12 & 53.83 \\
\ \ \ \ \ \textit{Iteration 2}   & 84.95 & 86.78 & 55.85 & 85.72 & 87.35 & 51.85 & 82.29 & 84.12 & 54.25 \\
\ \ \ \ \ \textit{Iteration 3}  & 84.78 & 86.72 & 55.63 & 85.58 & 87.33 & 51.85 & 82.20 & 84.08 & 54.45 \\ 
\rowcolor{mygray!60} \ \ \ \ \ \ \  \textit{+Distillation} & 83.86 & 86.21 & 54.88 & 84.75 & 86.68 & 50.98 & 81.47 & 83.50  & 53.73  \\
\midrule
SERM \\
\ \ \ \ \ \textit{Iteration 1}    & 87.04 & 87.53 & 56.50 & 87.72 & 88.12 & 52.74 & 84.55 & 84.96 & 54.24 \\
\ \ \ \ \ \textit{Iteration 2}    &  87.27 & 87.60 & 57.40 & 88.01 & 88.22 & 53.25 & 84.79 & 85.05 & 54.40 \\
\ \ \ \ \ \textit{Iteration 3}    & \bf87.56 & \bf87.71 & \bf57.79 & \bf88.14 & \bf88.27 & \bf53.51 & \bf84.99 & \bf85.12 & \bf55.07  \\ 
\rowcolor{mygray!60} \ \ \ \ \ \ \ \textit{+Distillation}  & 86.78 & 87.27 & 56.92 & 87.68 & 87.84 & 52.74 & 84.43 & 84.78 & 54.42 \\
\midrule
\multicolumn{5}{l}{\textbf{\textit{Training with Qwen2.5-1.5B Model}}}\\ \midrule
CT+SFT &   84.59 & 86.63 & 54.75 & 85.99 & 87.44 & 50.24 & 81.75 & 83.91 & 51.81  \\ 
\midrule
Self-Training  \\ 
\ \ \ \ \ \textit{Iteration 1}  & 84.91 & 86.77 & 55.22 & 86.07 & 87.51 & 51.08 & 82.00 & 84.02 & 53.09   \\
\ \ \ \ \ \textit{Iteration 2} & 84.93 & 86.75 & 55.49 & 85.98 & 87.47 & 51.22 & 82.10 & 84.05 & 53.32    \\
\ \ \ \ \ \textit{Iteration 3}  &  85.04 & 86.79 & 55.66 & 85.86 & 87.43 & 51.60 & 82.19 & 84.09 & 53.74 \\  
\rowcolor{mygray!60} \ \ \ \ \ \ \ \textit{+Distillation} &84.17 & 86.34 & 54.77 & 84.88 & 86.79 & 50.83 & 81.65 & 83.49 & 52.86   \\
\midrule
SERM    \\ 
\ \ \ \ \ \textit{Iteration 1}  & 86.64 & 87.38 & 55.57 & 87.46 & 88.02 & 51.98 & 84.11 & 84.81 & 53.51  \\
\ \ \ \ \ \textit{Iteration 2}  &  87.03 & 87.52 & 56.57 & 87.68 & 88.09 & 52.55 & 84.46 & 84.93 & 54.07 \\
\ \ \ \ \ \textit{Iteration 3}  &  \bf87.30 & \bf87.62 & \bf56.78 & \bf87.83 & \bf88.16 & \bf53.25 & \bf84.75 & \bf85.03 & \bf54.35 \\
\rowcolor{mygray!60} \ \ \ \ \ \ \ \textit{+Distillation} & 86.54 & 86.97 & 56.06 & 86.96 & 87.33 & 52.70  & 84.40  & 84.60  & 53.68   \\
\bottomrule[1.1pt]
\end{tabular}}
    \vspace{-2mm}
    \caption{
    Performance of relevance models on various language families. The best result in each group is in bold. ``+Distillation'' denotes the distillation of the relevance model from the third iteration to a small LLM (0.5B). ``ND@1'' and ``ND@4'' denote NDCG@1 and NDCG@4, respectively.
    }
    \vspace{-3mm}
    \label{tab:main_results}
\end{table*}

\subsection{Settings}
All of the trained relevance models, including the teacher and distilled models, employ a generative modeling approach, as described in Eq.~\ref{eq:generative_modeling}.
For the self-evolution phase, we executed three iterations, progressively augmenting the training dataset with each iteration. We used \texttt{GPT-4o} and \texttt{Gemini2.5-Pro} in the MRA module to generate three reasoning paths for each query–document pair.
During each iteration, we mixed the newly generated data, previously generated data, and the original SFT data, and retrained the model to prevent catastrophic forgetting \cite{wang2024hybrid,luo2025empirical}.
More training settings are shown in Appendix \ref{appendix:experimental_details}.

\subsection{Evaluation}
We conducted offline testing using an in-house test set that categorizes languages into three families: Germanic, Romance, and Minor Languages \cite{he2024scaling}, with the distribution shown in Table \ref{tab:data_dist}. This test set was specifically chosen to better reflect the language distribution and real-world application scenarios of our models.
We reported model performance based on NDCG@1, NDCG@4, and relevance accuracy (Acc.). 
Additionally, in industrial-scale search scenarios, relevance models must often be distilled into smaller models to meet strict latency requirements \cite{yao2022reprbert}.
Therefore, we also evaluated the performance of small models distilled from different trained relevance models. Specifically, we used Qwen2.5-0.5B as the small model and performed distillation using all of our SFT data, employing the Kullback-Leibler divergence-based distillation method as described in \citet{ye2025applying}'s work. 

\subsection{Baselines}
Our baseline for comparison was the traditional training pipeline, consisting of continual pre-training followed by SFT on static data (denoted as \textit{CT+SFT}). Additionally, to demonstrate the effectiveness of our multi-agent module, we compared it with the self-training approach. In this baseline, instead of using our MRA module for annotation, we relied on the relevance model to annotate the data itself (denoted as \textit{Self-Training}).
It is worth noting that during the self-training process, we ensured that the hyperparameters and query-document pairs used were consistent with those used in SERM to make a fair comparison.

\subsection{Offline Evaluation Results}
We conduct offline evaluations on the trained relevance models using a static, large-scale test set. The results are listed in Table \ref{tab:main_results}. First, compared to CT+SFT, we observe that both self-training and SERM show significant improvements, regardless of whether the model is Qwen2.5-7B or Qwen2.5-1.5B. This confirms that incorporating large volumes of user query streams into the training process effectively enhances model performance. Additionally, we observe that SERM outperforms self-training in terms of both stability and accuracy. This highlights the effectiveness of the multi-agent framework and the integration of external knowledge in improving model robustness. Interestingly, we find that noise can be easily introduced into the self-training process. For example, in the Qwen2.5-7B model, the performance of the third iteration on the Germanic language family significantly drops compared to the second iteration (e.g., NDCG@1 drops from 84.95 to 84.78). We conjecture that this is due to error propagation during the self-training iterations \cite{zhu2022meta}. In contrast, SERM consistently demonstrates stable improvements, which we attribute to the integration of external knowledge and annotations that mitigate such issues. Finally, when comparing the distilled models, we see that the model distilled from SERM Iteration 3 outperforms the one distilled from Self-Training Iteration 3. This further validates the effectiveness of the SERM approach in delivering superior relevance model performance.

\vspace{-2mm}

\subsection{Online Testing Results}

\paragraph{Online A/B Testing.}
We conduct an A/B test on the small model distilled from our 7B model (SERM Iteration 3), with the results summarized in Table \ref{tab:online_test}. From the results, we observe that our model can achieve a significant improvement in 14-day retention, with a +0.0359\% gain and a p-value of 0.0278, indicating enhanced long-term user engagement. We also find that the model can improve user satisfaction, as reflected by a slight decrease in user negative feedback by -1.2081\% (p-value: 0.0001)\footnote{On our platform, improvements of 0.01\% in metrics such as change query ratio and 14-day retention are considered significant due to the large user base (see Appendix~\ref{appendix:evaluation} for details). This is also consistent with prior work on industrial-scale evaluation \cite{zhou2025onerec}.}. These findings show the effectiveness of the SERM approach in enhancing both user retention and satisfaction through the self-evolution of the model using massive query streams.

\begin{table}[!t]
    \resizebox{\linewidth}{!}{
    \begin{tabular}{lcc}
\toprule[1.1pt]
Metric     & Gain      & P-value  \\ \hline
User Negative Feedback & -1.2081\% & 0.0001  \\
Change Query Ratio        & -0.0839\% & 0.0023  \\
Change Query Ratio (Longtail) & -0.1312\% & 0.0015  \\
14-Day Retention                 & +0.0359\% & 0.0278 \\ 
\bottomrule[1.1pt]
\end{tabular}
    }
    \vspace{-2mm}
    \caption{
    Performance metrics from an online A/B testing using the student model distilled from the Qwen2.5-7B model trained with SERM. The definitions of the metrics are provided in Appendix \ref{appendix:evaluation}.
    }
    \label{tab:online_test}
\end{table}

\begin{figure}[!t]
    \centering
    \definecolor{red}{RGB}{197,224,180} 
\definecolor{blue}{RGB}{157,195,230}

\begin{tikzpicture}
  \scriptsize{
  \begin{axis}[
    at={(-2.5em,0)},
    anchor=south west,
    ymajorgrids,
    grid style=dashed,
    legend style={at={(0.58,0.65)}, anchor=south west},
    legend cell align={right},
    ybar,
    enlarge x limits=0.2,
    xtick align=inside,
    height=.27\textwidth,
    width=.50\textwidth,
    bar width=2.em,
    xlabel={Language Family},
    xlabel style={scale=1.2, yshift=-0.5em},
    ylabel={$\Delta_{\mathrm{SBS}}$ (\%)},
    ylabel style={scale=1.2, yshift=0.3em},
    symbolic x coords={{1}, {2}, {3}},
    xtick=data,
    ymin=1.5,
    ymax=9.5,
    ytick={1,2,...,8,9},
    nodes near coords align={vertical},
    xticklabels={Germanic,Romance,Minor Language},
    x tick label style={
         anchor=center,
         scale=1.2,
         yshift=-0.8em
     },
    enlarge x limits=0.2,
    ylabel style={yshift=-2em,align=center},
    xlabel style={yshift=0.8em,align=center},
    yticklabel style={/pgf/number format/fixed,/pgf/number format/fixed zerofill,/pgf/number format/fixed,
    /pgf/number format/precision=1,rotate=0,scale=1.1},
    legend style={yshift=0.2em,xshift=2.5em,font={\tiny},cells={anchor=west},fill opacity=0.8, scale=0.2, legend columns=1},
    nodes near coords,
    nodes near coords style={font=\scriptsize, /pgf/number format/fixed, /pgf/number format/precision=1}
    ]
    \addplot[fill=red, draw=black, area legend] coordinates {({1},2.4) ({2},4.7) ({3},2.6) };
    \addlegendentry{\scalebox{1.2}{Random}}
    \addplot[fill=blue, draw=black, area legend] coordinates {({1},3.8) ({2},7.7) ({3},5.5) };
    \addlegendentry{\scalebox{1.2}{Longtail}}

  \end{axis}
  }
  
\end{tikzpicture}
    \vspace{-7mm}
    \caption{
    Side-by-side manual evaluation results comparing our proposed SERM with the baseline.
    }
    \vspace{-3mm}
    \label{fig:manual_evaluation}
\end{figure}

\paragraph{Side-by-Side Manual Evaluation.}
We further conduct a side-by-side (SBS) evaluation on our crowdsourcing platform, where annotators directly compare the outputs of the experimental system against the baseline to determine which provides better relevance. Detailed evaluation settings are described in Section~\ref{sec:side_by_side}. The advantage ratio of a given strategy is computed as:
\begin{eqnarray}
\Delta_{\mathrm{SBS}} = \frac{G - B}{G - B + S}
\end{eqnarray}
where $G$ denotes the number of cases in which the experimental strategy is preferred over the baseline, $B$ denotes the number of cases in which the baseline is preferred, and $S$ denotes the number of cases in which annotators see no clear difference (i.e., a tie). A higher $\Delta_{\mathrm{SBS}}$ indicates a stronger advantage of the experimental strategy over the baseline. The SBS results are summarized in Figure~\ref{fig:manual_evaluation}.
We can observe that SERM consistently outperforms the baseline on both random and long-tail test samples across all language families.

\subsection{Ablation Studies}
To evaluate the contribution of each agent within the MSM and MRA modules, we conduct an ablation study. Specifically, we re-run the SERM while removing one component at a time, including the user-feedback agent, click-model feedback agent, model disagreement, model uncertainty, inner-agreement agent, and inter-agreement agent. When the inner-agreement agent is removed, each LLM generates only a single reasoning path. When the inter-agreement agent is removed, we skip the external agreement filtering and instead directly aggregate all reasoning results from different LLMs via majority voting.

Table~\ref{tab:ablation_for_miner} summarizes the results. First, the results confirm that each component contributes valuable and complementary signals, enabling the agents to more effectively identify informative samples. The non-overlapping nature of these signals highlights the importance of integrating all components to achieve optimal self-evolution performance. Second, we find that both the inner-agreement and inter-agreement agents provide notable gains, underscoring their role in producing more reliable labels for the model’s iterative evolution. Interestingly, we see that although using a single GPT model alone for annotation can significantly degrade SERM’s performance due to its weaker capability on relevance tasks, GPT still proves beneficial for agreement analysis. This observation aligns with recent findings in \textit{weak-to-strong generalization} \cite{burns2023weak}, where a weak model can help a strong model improve further.

\begin{table}[!t]
    \centering
    \resizebox{\linewidth}{!}{
    \begin{tabular}{lccc}
\toprule[1.1pt]
Method & Germ. & Roma. & Minor.  \\  \midrule
CT+SFT  &86.68   &87.33   &84.02  \\ \midrule
SERM  &\textbf{87.71}   &\textbf{88.27}   &\textbf{85.12}  \\  
\ \ \ \ \ w/o User Feedback & 87.01 & 87.93 & 84.41  \\
\ \ \ \ \ w/o CM Feedback & 86.93 & 88.10 & 84.39  \\
\ \ \ \ \ w/o Model Disagreement & 86.95 & 87.74 & 84.57  \\
\ \ \ \ \ w/o Model Uncertainty &  87.17 & 87.51 & 84.69 \\ \hdashline
\ \ \ \ \ w/o Inner Agreement & 87.41 &	88.03 &	84.97   \\
\ \ \ \ \ w/o Inter Agreement (GPT) & 86.48 &	86.76 &	83.57   \\
\ \ \ \ \ w/o Inter Agreement (Gemini) & 86.84 & 87.47 & 84.36     \\

\bottomrule[1.1pt]
\end{tabular}}
    \vspace{-2mm}
    \caption{
    Ablation study on the MSM module of SERM using the Qwen2.5-7B model. 
    }
    \vspace{-4mm}
    \label{tab:ablation_for_miner}
\end{table}

\subsection{Effect of $\tau_{c}$ and $\tau_{\mathrm{cm}}$ on Performance.}
\begin{figure}[!t]
    \centering
    \vspace{1mm}
    \includegraphics[width=0.98\linewidth]{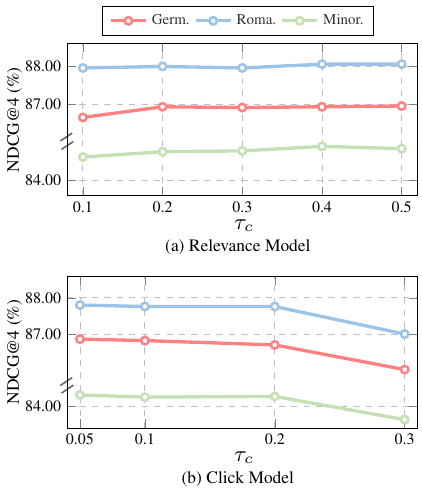}
    \vspace{-2mm}
    \caption{
    Performance on different thresholds.
    }
    \vspace{-4mm}
    \label{fig:diff_threshold}
\end{figure}

We further evaluate the performance of SERM under different hyperparameter settings.
Specifically, we conduct experiments using the Qwen2.5-7B model with varying thresholds for the relevance model ($\tau_{c}$) and the click model ($\tau_{\mathrm{cm}}$).
For $\tau_{c}$, we test values in $\{0.1, 0.2, 0.3, 0.4, 0.5\}$, and for $\tau_{\mathrm{cm}}$, we test values in $\{0.05, 0.1, 0.2, 0.3\}$.
We use a validation set with the same distribution as the test set to evaluate model performance under each configuration.
The results are presented in Figure~\ref{fig:diff_threshold}.
Based on these experiments, we select $\tau_{c} = 0.4$ and $\tau_{\mathrm{cm}} = 0.1$ as the default thresholds for SERM in subsequent self-evolution experiments.

\section{Related Work}
\paragraph{Relevance Modeling.}
Search relevance aims to rank items, such as documents, images, and videos, that best satisfy a user’s query intent. Early approaches relied on pre-trained encoders (e.g., BERT \cite{devlin2019bert}) and discriminative models \cite{gao2020understanding,zou2021pre,li2023parade,li2023pretrained}, while recent advances, inspired by the success of LLMs \cite{ouyang2022training,wang2024esrl,zhou2024prior,wang2025gram,wang2026probing}, have shifted toward generative approaches.
For instance, recent works have prompted LLMs to generate the relevance of either a single query–document pair or a query with multiple candidate documents \cite{qin2023large,zhuang2024setwise}. 
To further enhance LLM performance on search relevance tasks, some works have used labeled data to fine-tune the models, improving their ability to align with human-labeled relevance judgments \cite{ma2024fine}. Building on this, other works employed reinforcement learning to better exploit the reasoning capabilities of LLMs in relevance modeling \cite{tang2025lref,zhuang2025rank}.
More recently, to address the limitations of LLMs in capturing domain-specific patterns, researchers explored continual pre-training on large-scale search data \cite{zhang2023two,ye2025applying}. 

\paragraph{Self-Evolving Large Language Models.}
This work joins a large body of research demonstrating that LLMs can evolve themselves using unlabeled data \cite{huang2022large,wang2022self}.
One approach in this direction is \textit{self-training}, where the model leverages its own predictions on unlabeled data as pseudo-labels and iteratively retrains itself \cite{huang2022large,gulcehre2023reinforced}.
Despite its simplicity and effectiveness in semi-supervised and relatively stable settings, self-training often suffers from error propagation when the pseudo-labels are noisy or unreliable \cite{wang2021progressive,wang2026mro}.
Subsequent research has sought to address these limitations by enriching the learning signals beyond simple pseudo-labels \cite{madaan2023self,lu2023self}.
\vspace{-1mm}
\paragraph{Multi-Agent Collaboration.}
The emergence of LLMs has opened new possibilities for multi-agent collaboration \cite{zhu2025latextrans,tran2025multi}. In LLM-based multi-agent collaboration, each agent is instantiated as an LLM-powered entity capable of reasoning and planning \cite{qian2024scaling,guo2024large}. Systems such as AutoGPT,
CAMEL \cite{li2023camel}, and AutoGen \cite{wu2024autogen} demonstrate that LLM agents can assume diverse roles and accomplish complex tasks through dialogue-based coordination. In this work, we extend multi-agent collaboration to assist the self-evolution of relevance models.

\section{Conclusion}
We have explored how to learn from massive query streams to improve the generalization of relevance models. We have proposed a self-evolving relevance model approach, called SERM, which integrates a multi-agent sample miner and a multi-agent relevance annotator to enable continuous model adaptation.
We validated the effectiveness of SERM on a large-scale industrial document search platform through both offline and online testing. 
The experimental results show that SERM can continuously enhance model performance.

\section*{Limitations}
Search tasks are inherently diverse, encompassing not only document retrieval but also modalities such as image and video search. While the proposed SERM framework is applicable across these modalities, it is infeasible to exhaustively evaluate it on each task one by one. Therefore, this work primarily focuses on document search, which we consider a representative and widely adopted scenario across modern platforms.

\section*{Ethics Statement}
This work leverages large-scale real-world query streams, which naturally raise privacy and ethical concerns. We strictly follow anonymization and aggregation protocols and ensure that all data is legally obtained and free of harmful content. As a result, the work does not pose specific ethical risks.

\section*{Acknowledgments}
This work was supported in part by the National Natural Science Foundation of China (Nos. U24A20334 and 62276056), the Yunnan Fundamental Research Projects (No.202401BC070021), the Yunnan Science and Technology Major Project (No. 202502AD080014), the Fundamental Research Funds for the Central Universities (Nos. N25BSS054 and N25BSS094), and the Program of Introducing Talents of Discipline to Universities, Plan 111 (No.B16009). We would like to thank the anonymous reviewers and SPC for their valuable comments, which helped improve this paper.

\bibliography{custom}

\appendix

\clearpage


\section{Experimental Details}
\label{appendix:experimental_details}
\subsection{Setups}
\paragraph{Continual Pre-training.} 
Following prior work \cite{ye2025applying}, we first performed continual pre-training to help the model adapt to our specific experimental domain. During this stage, we set the learning rate to 1e-4 for both the Qwen2.5-7B and Qwen2.5-1.5B models.

\paragraph{SFT Training.}
During SFT, we used a learning rate of 3e-6 with a warm-up strategy applied during the first 10\% of the training process, and the training epoch was set to 1. In each iteration of SERM and self-training, we also adopted the same learning rate of 3e-6. Note that we applied identical hyperparameters for both the 7B and 1.5B models. We also experimented with adjusting the learning rate according to model size, but observed no significant improvement in performance. The prompt used for SFT training is shown in Figure~\ref{fig:prompt_for_generative_model}.
For each input, the document included its title, hashtags, and summary. 

\paragraph{SERM.}
In our experiments with SERM, we configured the agents as follows.
For the user-feedback agent, we set the dwell-time threshold $\tau_{u}$ to 5 seconds and the model confidence threshold $\tau_{c}$ to 0.4.
Note that $\tau_{u}$ was determined based on platform-specific statistics derived from real-world business scenarios, while $\tau_{c}$ was selected empirically as the optimal value, as shown in Figure~\ref{fig:diff_threshold}. During sample selection, we set $n$ to 4, meaning that for each query, the sample-mining agents collectively select up to four candidate documents to form the training samples.
If the number of documents satisfying the agent’s selection criteria exceeds $n$, we randomly sample four; if it is fewer than $n$, we use all qualifying documents.
For the click-model feedback agent, we set the threshold $\tau_{\mathrm{cm}}$ to 0.1.
For the inner-agreement agent, we generated three reasoning paths for each query–document pair to improve label stability and consistency. The prompt used for this process is provided in Figure~\ref{fig:llm_annotation_prompt}.

\begin{figure}[!t]
    \centering
    \definecolor{red}{RGB}{197,224,180} 
\definecolor{blue}{RGB}{157,195,230}

\begin{tikzpicture}[font=\scriptsize]
  \begin{axis}[
    at={(-2.5em,0)},          
    anchor=south west,
    ymajorgrids,
    grid style=dashed,
    legend style={at={(0.58,0.65)}, anchor=south west},
    legend cell align={right},
    ybar,
    enlarge x limits=0.2,
    xtick align=inside,
    height=.28\textwidth,
    width=.25\textwidth,      
    bar width=1.em,
    xlabel={Length Range},    
    xlabel style={scale=1.2,yshift=-0.9em, align=center},
    ylabel={Frequency (\%)},  
    ylabel style={scale=1.2, yshift=-0.5em, align=center},
    symbolic x coords={{1}, {2}, {3}, {4}}, 
    xtick=data,
    ymin=0,                   
    ymax=61,                  
    ytick={0,10,20,30,40,50,60}, 
    nodes near coords align={vertical},
    xticklabels={{[0,50]}, {(50,100]}, {(100,200]}, {(200,1000]}},
    x tick label style={
         anchor=center,
         scale=1.0,           
         yshift=-1.2em,
         rotate=45
     },
    yticklabel style={
      /pgf/number format/fixed,
      /pgf/number format/fixed zerofill,
      /pgf/number format/precision=1,
      rotate=0,
      scale=1.1
    },
    legend style={
      yshift=0.2em,
      xshift=2.5em,
      font={\tiny},
      cells={anchor=west},
      fill opacity=0.8, 
      scale=0.2, 
      legend columns=1
    },
    nodes near coords,
    nodes near coords style={
      font=\scriptsize, 
      /pgf/number format/fixed, 
      /pgf/number format/precision=1
    },
    title={Pre-Training},   
    title style={scale=1.1, yshift=-0.3em}
  ]
  \addplot[fill=red, draw=black, area legend] coordinates {
    (1,0.39)
    (2,43.6)
    (3,52.4)
    (4,3.6)
  };
  \end{axis}

  \begin{axis}[
    at={(0.18\textwidth,0)},   
    anchor=south west,
    ymajorgrids,
    grid style=dashed,
    legend style={at={(0.58,0.65)}, anchor=south west},
    legend cell align={right},
    ybar,
    enlarge x limits=0.2,
    xtick align=inside,
    height=.28\textwidth,
    width=.25\textwidth,      
    bar width=1.em,
    xlabel={Length Range},    
    xlabel style={scale=1.2, yshift=-0.9em, align=center},
    ylabel={Frequency (\%)},  
    ylabel style={scale=1.2, yshift=-0.5em, align=center},
    symbolic x coords={{1}, {2}, {3}, {4}}, 
    xtick=data,
    ymin=0,                   
    ymax=95,                  
    ytick={0,20,40,60,80},    
    nodes near coords align={vertical},
    xticklabels={{[0,10]}, {(10,20]}, {(20,30]}, {(30,1000]}},
    x tick label style={
         anchor=center,
         scale=1.0,           
         yshift=-1.2em,
         rotate=45
     },
    yticklabel style={
      /pgf/number format/fixed,
      /pgf/number format/fixed zerofill,
      /pgf/number format/precision=1,
      rotate=0,
      scale=1.1
    },
    legend style={
      yshift=0.2em,
      xshift=2.5em,
      font={\tiny},
      cells={anchor=west},
      fill opacity=0.8, 
      scale=0.2, 
      legend columns=1
    },
    nodes near coords,
    nodes near coords style={
      font=\scriptsize, 
      /pgf/number format/fixed, 
      /pgf/number format/precision=2 
    },
    title={SFT},     
    title style={scale=1.1, yshift=-0.3em}
  ]
  \addplot[fill=blue, draw=black, area legend] coordinates {
    (1,82.4)
    (2,17.4)
    (3,0.2)
    (4,0.01)
  };
  \end{axis}
\end{tikzpicture}
    \vspace{-8mm}
    \caption{Length distribution of document datasets and queries used in our experiments.}
    \label{fig:data_dist_2}
\end{figure}

\begin{table}[!t]
    \centering
    \resizebox{\linewidth}{!}{
    
\begin{tabular}{lccc}
\toprule[1.1pt]
\multirow{2}{*}{Dataset}    & Germanic & Romance     & Minor Language \\  \cmidrule(l){2-2} \cmidrule(l){3-3} \cmidrule(l){4-4}
& EN DE NL & ES PT FR IT & ID AR JA RU    \\ \midrule
 SFT  & 1,655,040  & 1,105,927     & 840,963         \\ \midrule
SERM (Iteration 1)  & 359,249   &190,630      & 140,123         \\
SERM (Iteration 2) & 354,551   & 172,343  & 159,978         \\
SERM (Iteration 3)  & 358,090   & 183,872  & 160,692         \\ 
SERM (Iteration 4)  & 328,506   & 177,393  & 144,542       \\ 
SERM (Iteration 5)  & 335,001   & 174,200 & 160,800        \\ 
\midrule
Testing     &43,793    &23,392       &18,262          \\    
Testing-v2  &57,126    &33,391       &30,261         \\    

\bottomrule[1.1pt]
\end{tabular}}
    \caption{
    The language distribution and statistics for the training and test datasets, including data from SFT and self-evolution (SERM), are based on different language families: Germanic, Romance, and Minor Languages, following the classification of \citet{he2024scaling}. Note that ``Testing-v2'' denotes our latest testing set, which incorporates newly collected test cases to evaluate better whether the model continues to self-evolve over time. The construction and curation of the testing-v2 dataset are described in Appendix~\ref{sec:results_from_additional_serm}. 
    }
    \vspace{-3mm}
    \label{tab:data_dist}
\end{table}

\subsection{Datasets}
We present the length distributions of the datasets used in the pre-training and SFT stages in Figure~\ref{fig:data_dist_2}. Both the continual pre-training and SFT pipelines follow standard industry practices and do not include any procedures specifically designed to favor our SERM. Moreover, the self-training baseline is constructed using the same pre-training and SFT setup, which effectively isolates the impact of these stages from the performance gains observed in our experiments.
As shown in Table~\ref{tab:main_results}, under an identical training recipe with the Qwen2.5-7B model, SERM consistently outperforms the self-training baseline by 1--2 points in NDCG@4 at Iteration~3. This result demonstrates that the improvements achieved by SERM are not attributable to differences in pre-training or SFT data, but rather stem from the proposed self-evolving framework itself.
Additionally, we present the language distribution and dataset statistics for both the training and test sets in Table~\ref{tab:data_dist}. Our evaluation is conducted in a multilingual search environment, which poses additional challenges and further highlights the robustness of the proposed approach.

\begin{table*}[!t]
\centering
\resizebox{1.0\linewidth}{!}{
\begin{tabular}{lccc}
\toprule
\textbf{Metric} & \textbf{Absolute Gain} & \textbf{P-value} & \textbf{Unit} \\
\midrule
User Negative Feedback         & -2,416,200 & 0.0001 & counts (number of events) \\
Change Query Ratio             & -2,409,524 & 0.0023 & counts (number of change-query events) \\
Change Query Ratio (Longtail)  &-753,586 & 0.0015 & counts (number of change-query events) \\
14-Day Retention               & +25,130 & 0.0278 & user-days \\
\bottomrule
\end{tabular}
}
\caption{
Absolute improvements observed in the online A/B test. 
}
\vspace{-2mm}
\label{tab:ab_absolute_gain}
\end{table*}

\vspace{-2mm}
\subsection{Evaluation}
\label{appendix:evaluation}
\paragraph{Inference of Generative Relevance Models.}
In this work, we employ a probability aggregation approach to derive relevance scores from the generative relevance model.
Specifically, given a query–document pair $(q, d)$, the model generates a discrete relevance label token (e.g., $y \in \{0,1,2,3,4\}$), where the predicted token reflects the model’s relevance judgment.
To obtain a continuous relevance score that is more informative for downstream ranking, we compute the expectation over the label probabilities as:
\begin{eqnarray}
f(q,d) = \sum_{y \in \mathcal{Y}} y \cdot \mathrm{Pr}_\theta\big(y \mid q,d\big)
\label{eq:obtain_score_from_grm}
\end{eqnarray}
where $\mathrm{Pr}_\theta(y \mid q,d)$ denotes the probability of generating label $y$ under the model parameters $\theta$.

\paragraph{A/B Testing Metrics.}
We report the following key metrics for evaluating the impact of relevance models in online A/B testing:
\begin{itemize}
    \item \textit{User Negative Feedback}. This metric captures instances where users provide explicit negative feedback (e.g., reporting irrelevant or unsatisfactory results). A lower value indicates higher immediate user satisfaction.
    \item \textit{Change Query Ratio}. This metric measures the proportion of cases in which users reformulate or issue a new query for the same information need after the initial search did not meet their expectations. A lower ratio suggests that the system is more effective at fulfilling user intent on the first attempt.
    \item \textit{14-Day Retention}. This metric tracks whether users continue to return to the platform over a 14-day window. A higher value indicates stronger long-term engagement and improved user loyalty.
\end{itemize}
In the experimental results, although the absolute percentage improvement appears small, the practical impact is, in fact, substantial due to the extremely large user base involved in our online A/B evaluation. 
For instance, both the control and treatment groups contain approximately 70 million users. Under such a scale, even a 0.0359\% improvement translates to a significant increase in cumulative user activity:
\begin{eqnarray}
0.000359 \times 70{,}000{,}000  = 25{,}130 \ \text{user-days}
\end{eqnarray}
We also present all the absolute improvements observed in our A/B test results in Table \ref{tab:ab_absolute_gain}. From the results, although the percentage appears small, the gain is far from negligible in practice; rather, it represents a meaningful improvement in long-term user engagement within a high-traffic industrial search system. It is also worth noting that such A/B online testing metrics are commonly used in search relevance evaluation, and the corresponding improvements are typically small in absolute terms due to the extremely large user bases involved \cite{zou2021pre,ye2025applying,zhou2025onerec}.

\paragraph{Side-by-Side Evaluation.}
\label{sec:side_by_side}
In our side-by-side evaluation, the random category refers to a uniformly sampled subset of user queries drawn from the overall production traffic, reflecting the average distribution of common and frequently occurring queries. In contrast, the long-tail category consists of queries that fall into the low-frequency region of the query distribution, \textit{i.e.}, those appearing below a frequency threshold within the same logging period. These long-tail queries are typically rare, emerging, or domain-specific, and are underrepresented in training data, making them substantially more challenging for relevance models to handle. Prior work in information retrieval has similarly emphasized the importance of evaluating models on low-frequency or long-tail queries, as they better reflect robustness and generalization in practical deployments \cite{ye2025applying}.

\begin{table*}[!t]
    \centering
    \resizebox{\linewidth}{!}{
    \setlength{\tabcolsep}{6.5pt}
\begin{tabular}{lccccccccc}
\toprule[1.1pt]
\multirow{2}{*}{Method} &   \multicolumn{3}{c}{Germanic} & \multicolumn{3}{c}{Romance} & \multicolumn{3}{c}{Minor Language} \\ \cmidrule(r){2-4} \cmidrule(r){5-7} \cmidrule(r){8-10}
&ND@1 & ND@4  & Acc.  & ND@1     &  ND@4  &Acc.  &ND@1 &ND@4  &Acc.  \\ \midrule
\multicolumn{5}{l}{\textbf{\textit{Training with Qwen2.5-7B Model}}}\\ \midrule
CT+SFT & 84.74 & 86.68 & 55.30 & 85.61 & 87.33 & 50.74 & 82.02 & 84.02 & 52.89  \\ 
\midrule
Self-Training  \\
\ \ \ \ \ \textit{Iteration 4}  & 84.55     & 86.63     & 55.54   & 85.44     & 87.26     & 51.80    & 82.07     & 84.03     & 54.34\\
\ \ \ \ \ \textit{Iteration 5}   &  84.44     & 86.60      & 55.53   & 85.42     & 87.27     & 51.80    & 82.01     & 84.02     & 54.34  \\
\midrule
SERM \\
\ \ \ \ \ \textit{Iteration 4}    & \textbf{87.64}     & \textbf{87.74}     & \textbf{57.90}    & \textbf{88.04}     & \textbf{88.19}     & \textbf{53.96}   & \textbf{84.98}     & \textbf{85.12}     & 55.49 \\
\ \ \ \ \ \textit{Iteration 5}    &  87.47     & 87.68     & 57.68   & 87.86     & 88.18     & 53.46   & 84.89     & 85.08     & \textbf{55.69} \\
\midrule
\multicolumn{5}{l}{\textbf{\textit{Training with Qwen2.5-1.5B Model}}}\\ \midrule
CT+SFT &   84.59 & 86.63 & 54.75 & 85.99 & 87.44 & 50.24 & 81.75 & 83.91 & 51.81  \\ 
\midrule
Self-Training  \\ 
\ \ \ \ \ \textit{Iteration 4}  & 84.63     & 86.65     & 55.62   & 85.31     & 87.23     & 51.55   & 81.86     & 83.97     & 53.70   \\
\ \ \ \ \ \textit{Iteration 5} &  84.70      & 86.66     & 55.63   & 85.46     & 87.26     & 51.57   & 81.80      & 83.96     & 53.70   \\
\midrule
SERM    \\ 
\ \ \ \ \ \textit{Iteration 4}  & 87.32     & 87.60      & 57.05   & \textbf{87.74}     & \textbf{88.12}     & 53.39   & 84.85     & 85.06     & 54.58  \\
\ \ \ \ \ \textit{Iteration 5}  & \textbf{87.43}     & \textbf{87.64}  & \textbf{57.22}   & 87.62     & 88.08     & \textbf{53.77}   & \textbf{84.94}     & \textbf{85.10}    & \textbf{55.00}  \\
\bottomrule[1.1pt]
\end{tabular}
    }    
    \caption{
    Results from additional SERM iterations 4 and 5. 
    }
    \label{tab:additional_iteratives}
    \vspace{-2mm}
\end{table*}

\section{More Analysis}
\subsection{Differences from Knowledge Distillation}
\label{app:diff_from_kd}
In this subsection, we discuss how SERM differs from conventional knowledge distillation in two key aspects. First, \textit{unlike traditional knowledge distillation where a student passively mimics a static teacher on a fixed dataset, SERM acts as an active learner}. The ``evolution'' is driven by the model’s own internal states. Specifically, the MSM uses the model’s own uncertainty and disagreement (as shown in Section \ref{sec:intrinsic_feedback_agent}) to identify where it is failing. In this way, the model itself dictates the curriculum of its learning process. If the model were confident and correct, no evolution would be triggered. Thus, the impetus for improvement is intrinsic, even if the supervision is extrinsic. Second, \textit{while traditional knowledge distillation operates in a static manner, where a fixed teacher transfers knowledge to a student before deployment, our SERM presents a fundamentally different paradigm}. Specifically, unlike knowledge distillation, SERM remains adaptive after deployment and is driven by three forms of dynamism that are intrinsic to industrial search environments: a dynamic environment, dynamic data, and dynamic annotation.

\subsection{Results from Additional SERM Iterations}
\label{sec:results_from_additional_serm}

\paragraph{Performance Saturation on the Static Testing Set.} 
Table~\ref{tab:additional_iteratives} reports results from SERM iterations 4 and 5 trained on newly collected queries from the most recent month. We observe that performance gains gradually saturate across later iterations, particularly for the self-training baseline, whose metrics remain nearly unchanged from iteration 4 to iteration 5. This suggests that, on this relatively static test distribution, the model has approached the upper bound of achievable performance. Notably, despite this saturation effect, SERM consistently outperforms self-training across all language families and both model scales. Even in later iterations, SERM maintains stable improvements of approximately 1–2 points in NDCG@4 and accuracy, indicating that it continues to provide higher-quality evolutionary signals than standard self-training.  These observations naturally raise an important question: \textit{Does the observed saturation indicate that SERM has reached its capacity for further improving relevance models?}

\begin{table*}[!t]
    \centering
    \resizebox{\linewidth}{!}{
    \setlength{\tabcolsep}{6.5pt}
\begin{tabular}{lccccccccc}
\toprule[1.1pt]
\multirow{2}{*}{Method} &   \multicolumn{3}{c}{Germanic} & \multicolumn{3}{c}{Romance} & \multicolumn{3}{c}{Minor Language} \\ \cmidrule(r){2-4} \cmidrule(r){5-7} \cmidrule(r){8-10}
&ND@1 & ND@4  & Acc.  & ND@1     &  ND@4  &Acc.  &ND@1 &ND@4  &Acc.  \\ \midrule
\multicolumn{5}{l}{\textbf{\textit{Training with Qwen2.5-7B Model}}}\\ \midrule
CT+SFT &  71.69 & 71.29 & 47.22 & 73.94 & 73.30 & 48.94 & 70.08 & 70.17 & 47.42 \\ 
\midrule
Self-Training  \\
\ \ \ \ \ \textit{Iteration 1} & 72.07     & 71.76     & 47.75   & 74.02     & 73.45     & 49.03   & 70.29     & 70.51     & 47.56  \\
\ \ \ \ \ \textit{Iteration 2} & 72.22     & 72.31     & 47.82   & 74.33     & 73.77     & 49.31   & 70.36     & 70.68     & 47.65 \\
\ \ \ \ \ \textit{Iteration 3} & 72.31     & 71.99     & 48.25   & 74.19     & 73.54     & 49.43   & 70.50     & 70.69     & 48.04 \\
\ \ \ \ \ \textit{Iteration 4} & 72.46     & 72.44     & 48.44   & 74.31     & 73.93     & 50.02   & 70.76     & 70.87     & 48.06 \\
\ \ \ \ \ \textit{Iteration 5}  & 72.64     & 72.14     & 48.47   & 74.37     & 73.60     & 50.12   & 70.74     & 70.80     & 48.50 \\
\midrule
SERM \\
\ \ \ \ \ \textit{Iteration 1} & 74.07     & 73.02     & 49.96   & 75.62     & 75.29     & 50.62   & 74.11     & 74.17     & 50.48 \\
\ \ \ \ \ \textit{Iteration 2} & 75.42     & 74.10      & 50.59   & 75.98     & 76.87     & 50.69   & 74.85     & 74.41     & 50.65 \\
\ \ \ \ \ \textit{Iteration 3} & 75.43     & 74.38     & 50.84   & 76.58     & 77.51     & 50.91   & 75.10      & 74.48     & 50.61  \\
\ \ \ \ \ \textit{Iteration 4} & 76.21     & 74.47     & 51.15   & 77.64     & 77.52     & 53.54   & 76.51     & 75.34     & 52.76 \\
\ \ \ \ \ \textit{Iteration 5} & \textbf{76.67}     & \textbf{74.57}     & \textbf{51.33}   & \textbf{78.04}     & \textbf{77.76}     & \textbf{53.56}   & \textbf{76.63}     & \textbf{75.88}     & \textbf{53.28} \\
\midrule
\multicolumn{5}{l}{\textbf{\textit{Training with Qwen2.5-1.5B Model}}}\\ \midrule
CT+SFT & 71.33 & 70.94 & 46.65 & 73.72 & 72.54 & 48.69 & 68.90 & 70.09 & 46.53    \\ 
\midrule
Self-Training  \\ 
\ \ \ \ \ \textit{Iteration 1} & 71.94     & 71.19     & 47.44   & 73.84     & 72.80     & 48.84   & 69.50     & 70.48     & 47.36 \\
\ \ \ \ \ \textit{Iteration 2} & 72.15     & 72.49     & 47.52   & 74.11     & 73.01     & 48.62   & 69.97     & 70.75     & 47.45 \\
\ \ \ \ \ \textit{Iteration 3} & 72.17     & 71.36     & 47.76   & 73.92     & 72.96     & 49.42   & 69.67     & 70.75     & 47.52 \\
\ \ \ \ \ \textit{Iteration 4} & 72.26     & 72.66     & 47.53   & 74.06     & 73.18     & 48.83   & 70.16     & 70.87     & 48.25  \\
\ \ \ \ \ \textit{Iteration 5} & 72.38     & 71.49     & 48.14   & 74.13     & 73.02     & 49.70   & 70.02     & 70.88     & 48.01  \\
\midrule
SERM    \\ 
\ \ \ \ \ \textit{Iteration 1} & 73.94     & 73.19     & 48.70    & 74.55     & 76.75     & 50.95   & 72.97     & 72.95     & 49.86  \\
\ \ \ \ \ \textit{Iteration 2} & 74.25     & 73.49     & 49.06   & 74.68     & 76.78     & 51.51   & 73.04     & 72.98     & 49.96 \\
\ \ \ \ \ \textit{Iteration 3} & 74.49     & 73.78     & 49.28   & 76.30      & 76.96     & 51.63   & 73.09     & 73.16     & 50.12  \\
\ \ \ \ \ \textit{Iteration 4} & 74.83     & 73.82     & 49.87   & 76.81     & 77.04     & 52.44   & 73.31     & 74.00        & 51.08  \\
\ \ \ \ \ \textit{Iteration 5} & \textbf{75.24}     & \textbf{73.83}     & \textbf{50.32}   & \textbf{77.49}     & \textbf{77.09}     & \textbf{52.75}   & \textbf{73.81}     & \textbf{74.30}      & \textbf{51.59} \\
\bottomrule[1.1pt]
\end{tabular}
    }
    \caption{
    Results on a newly collected testing set with evolving query distributions.
    }
    \vspace{-2mm}
    \label{tab:results_on_new_testingset}
\end{table*}

\vspace{-1mm}
\paragraph{Results on a Newly Collected Testing Set.}
To further investigate this question, we construct a newly collected testing set, denoted as \textit{testing-v2}, which incorporates more recent and evolving user queries. Specifically, testing-v2 augments the original testing set with newly emerged query cases collected in the most recent month, aiming to better reflect the continuously shifting query distribution in real-world search environments. The language distribution and statistics of the testing-v2 are presented in Table~\ref{tab:data_dist}.
Note that these additional test queries are not selected based on or aligned with the training data used in Iterations 4 and 5, ensuring a fair and unbiased evaluation.
We re-evaluate models from Iterations 1 to 5 on testing-v2, and the results are reported in Table~\ref{tab:results_on_new_testingset}. Several important observations can be drawn. First, across all language families and both model scales, SERM consistently outperforms the self-training baseline at every iteration, reaffirming the robustness and general effectiveness of the proposed self-evolution framework under distribution shifts. 
Second, unlike the saturation behavior observed in Table~\ref{tab:additional_iteratives}, SERM continues to exhibit monotonic or near-monotonic improvements across iterations on testing-v2, particularly in later iterations. This contrast suggests that the previously observed performance plateau is largely attributable to the limited capacity of a fixed test distribution, rather than a degradation or failure of the SERM mechanism itself. In other words, while the model may appear saturated under static evaluation, it continues to acquire new capabilities when assessed against evolving query distributions.


\clearpage

\begin{figure*}[!t]
    \centering
    \resizebox{\linewidth}{!}{
    \input{images/prompt_training_generative_model}
    }
    \caption{
    Template used for training our generative relevance models.
    }
    \label{fig:prompt_for_generative_model}
\end{figure*}

\clearpage

\begin{figure*}
    \centering
    \resizebox{\linewidth}{!}{
    \input{images/prompt_llm_labeling}}
    \caption{
    Template used for labeling relevance judgments in the MRA module.
    }
    \label{fig:llm_annotation_prompt}
\end{figure*}

\clearpage

\begin{table*}
    \centering
    \begin{tabular}{ll}
\toprule[1.1pt]
\bf{Query} &  
\parbox{0.77\linewidth}{
tongue piercing
} \\ \midrule
\multicolumn{2}{l}{\textbf{Document}}   \\ 
Title   &
\parbox{0.77\linewidth}{
 I’m low-key crying
}     \\ \hdashline
Summary    & 
\parbox{0.77\linewidth}{
Hey, it's been almost three months since I got my tongue piercing. I changed it today like I have a few times before, and it was bothering me. I thought it was because of the new piercing jewelry I put in, but now I can’t get it back in at all. Good afternoon. Come by this afternoon and we’ll take a look. Bring all the jewelry pieces you have, or any that you know fit you well. What time do you close? We’re open until 8:15 PM.
}      \\ \hdashline
Hashtag  & 
\parbox{0.77\linewidth}{
    \#crying \#tonguepiercing 
}      \\ \midrule
\midrule
\multicolumn{2}{l}{\textbf{LLM Reasoning \#1}}   \\  
Rationale    & 
\parbox{0.77\linewidth}{
The document discusses the author's experience with a tongue piercing, including issues with changing the piercing and seeking advice. This is somewhat relevant to the query `tongue piercing' as it provides personal insights and experiences related to tongue piercings, but it does not fully satisfy the query as it lacks detailed or instructional content.
}      \\ \hdashline
Relevance Score    & 
\parbox{0.77\linewidth}{
    2
}      \\ \midrule
\multicolumn{2}{l}{\textbf{LLM Reasoning \#2}}   \\  
Rationale & \parbox{0.77\linewidth}{
The document content is related to the query `tongue piercing' because it talks about a tongue piercing and the author's experience with it. However, it does not provide comprehensive or highly useful information, such as how to get a tongue piercing or care tips, which limits its relevance.
}      \\ \hdashline
Relevance Score    & 
\parbox{0.77\linewidth}{
    2
}      
\\ \midrule
\multicolumn{2}{l}{\textbf{LLM Reasoning \#3}}   \\  
Rationale   & 
\parbox{0.77\linewidth}{
While the document mentions a tongue piercing and some issues the author faced, it does not provide substantial or actionable information about tongue piercings. It is somewhat relevant but not highly valuable for someone searching for `tongue piercing'.
}      
\\ \hdashline
Relevance Score    & 
\parbox{0.77\linewidth}{
    1
}      \\ \bottomrule[1.1pt]
\end{tabular}
    \caption{
    We present a case where the LLM generates reasoning and relevance scores for a query–document pair in the MRA module. Three different reasoning paths are provided for the query ``tongue piercing'' and a document that records a conversation about the author's experience with a tongue piercing issue and seeking related advice.
    }
    \label{tab:case_for_llm_annotation}
\end{table*}

\clearpage

\begin{table*}[!t]
    \centering
    \begin{tabular}{ll}
\toprule[1.1pt]
\bf{Query} &  
\parbox{0.77\linewidth}{
wild au   aughhh
} \\ \midrule
\multicolumn{2}{l}{\textbf{Document}}   \\ 
Title   &
\parbox{0.77\linewidth}{
-wild au-
}     \\ \hdashline
Summary    & 
\parbox{0.77\linewidth}{
cupcakewild au- (long intro) intro:10sec magbago kana uy
}      \\ \hdashline
Hashtag  & 
\parbox{0.77\linewidth}{
\#aufornow \#enhypenau \#aumeaning \#wildau \#yoyoyoyoyoyo
}      \\ \midrule
\midrule
\multicolumn{2}{l}{\textbf{Human-Annotated Rationale}} \\
\multicolumn{2}{l}{
\parbox{0.96\linewidth}{
Analyzing the query `wild au aughhh', the most important keywords are `wild' and `au'. The document's title `-wild au-' and the hashtag `\#wildau' directly correspond to these keywords. The term `au' in the context of TikTok and hashtags like `\#enhypenau' often refers to `Alternate Universe', a popular fanfiction genre. So, the user is likely searching for a `wild' themed `Alternate Universe' story or document. This document is explicitly labeled as a `wild au'. The other hashtags like `\#aufornow' and `\#aumeaning' further confirm the content is about the `au' concept. The document is highly relevant and directly addresses the user's search intent.
}}      
\\ \midrule
CT+SFT   & 
\parbox{0.77\linewidth}{
    0.6193
}      \\ \hdashline 
Self-Training   & 
\parbox{0.77\linewidth}{
    0.7866 (Iteration 1) $\rightarrow$ 0.6586 (Iteration 2)$\rightarrow$ 0.4854 (Iteration 3)
}      \\ \hdashline 
SERM    & 
\parbox{0.77\linewidth}{
    0.7378 (Iteration 1) $\rightarrow$ 0.8735 (Iteration 2)$\rightarrow$ 0.9389 (Iteration 3)
}      \\  \bottomrule[1.1pt]
\end{tabular}
    \caption{
    Case 1 of relevance scores assigned by different models. In contrast to Self-Training, which suffers from performance degradation across iterations, SERM progressively captures the true intent behind the query (``wild au aughhh'') and assigns increasingly accurate relevance scores to the document, demonstrating superior query understanding and more effective relevance modeling.
    }
    \label{tab:case1}
\end{table*}

\begin{table*}[!t]
    \centering
    \begin{tabular}{ll}
\toprule[1.1pt]
\bf{Query} &  
\parbox{0.77\linewidth}{
how to  prepare garlic and lemon
} \\ \midrule
\multicolumn{2}{l}{\textbf{Document}}   \\ 
Title   &
\parbox{0.77\linewidth}{
Lemon+Garlic
}     \\ \hdashline
Summary    & 
\parbox{0.77\linewidth}{
Lemon + Garlic. Mix lemon with garlic and you will not visit a hospital again. A single cup of this drink is enough to clean your kidneys, skin, and much more. For this powerful drink, we will use a previously washed lemon. Lemon provides a large amount of vitamin C, potassium, and smaller amounts of other vitamins and minerals. Finally, we will use a piece of turmeric.
}      \\ \hdashline
Hashtag  & 
\parbox{0.77\linewidth}{
\#powerofherbs \#halamanggamot \#herbalplants \#healthbenefits}      \\ \midrule
\midrule
\multicolumn{2}{l}{\textbf{Human-Annotated Rationale}}   \\  
\multicolumn{2}{l}{
\parbox{0.96\linewidth}{
The user's query is `how to prepare garlic and lemon'. The document title is `Lemon+Garlic' and the summary explicitly mentions mixing lemon with garlic. The summary starts by saying `Mix lemon with garlic...' and then lists the ingredients needed: `a previously washed lemon' and 'a piece of turmeric'. Although it doesn't give the exact step-by-step preparation instructions (like chopping, blending, quantities), it clearly indicates that it's a recipe or preparation guide. The document is about preparing a drink with lemon and garlic, which directly aligns with the user's intent to learn how to prepare these two ingredients together. Therefore, the document is highly relevant and likely contains the exact preparation method the user is looking for.
}}      \\ \midrule
CT+SFT   & 
\parbox{0.77\linewidth}{
    0.7429
}      \\ \hdashline 
Self-Training   & 
\parbox{0.77\linewidth}{
    0.7371 (Iteration 1)  $\rightarrow$ 0.7059 (Iteration 2) $\rightarrow$ 0.9675 (Iteration 3) 
}      \\ \hdashline 
SERM    & 
\parbox{0.77\linewidth}{
    0.9384 (Iteration 1)  $\rightarrow$ 0.9743 (Iteration 2) $\rightarrow$ 0.9839 (Iteration 3) 
}      \\  \bottomrule[1.1pt]
\end{tabular}
    \caption{
    Case 2 of relevance scores assigned by different models. As shown, SERM increasingly assigns higher relevance scores across iterations, demonstrating its ability to capture and align with the query’s true intent—identifying that the document provides guidance on preparing a drink with lemon and garlic. 
    }
    \label{tab:case1}
\end{table*}

\end{document}